\documentclass[acmlarge]{acmart}

\usepackage{booktabs} 

\usepackage[ruled]{algorithm2e} 
\SetAlFnt{\small}
\SetAlCapFnt{\small}
\SetAlCapNameFnt{\small}
\SetAlCapHSkip{0pt}
\IncMargin{-\parindent}

\begin{document}
\title{AI Meets Physical World -- Exploring Robot Cooking}

\author{Yu Sun}
\affiliation{%
  \institution{Robot Perception and Action Lab, Department of Computer Science and Engineering, University of South Florida}
  \city{Tampa}
  \state{FL}
  \postcode{33620}
  \country{USA}}
\email{yusun@usf.edu}

\begin{abstract}
This paper describes our recent research effort to bring the computer intelligence into the physical world so that robots could perform physically interactive manipulation tasks. Our proposed approach first gives robots the ability to learn manipulation skills by ``watching'' online instructional videos. After ``watching'' over 200 instructional videos, a functional object-oriented network (FOON) is constructed to represent the observed manipulation skills. Using the network, robots can take a high-level task command such as ``I want BBQ Ribs for dinner,'' decipher the task goal, seek the correct objects to operate on, and then generate and execute a sequence of manipulation motions that can deal with physical interactions in a new condition. To best facilitate manipulation motions in the physical world, we also developed new grasping strategies for robots to hold objects with a firm grasp to withstand the disturbance during physical interactions. 
\end{abstract}

%


%
%



\maketitle


\section{Introduction}
Computers have tremendous learning capability and have demonstrated superior intelligence in many domains. However, it has yet to transfer the success in the cyber world (data mining, gaming, etc.) into the physical world, so that robots could help us in daily-living tasks. For example, robot cooking would be very useful especially for seniors and people with severe physical disabilities. Since cooking involves many manipulation tasks that requires physical interaction, such as cutting, it is still very challenging for a robot to perform those tasks in a general kitchen setting. 

Programming a robot to perform cooking task-by-task doesn't scale well in daily-living environment. Recent researches have dramatically improved the robot's capability and flexibility along the progressing of computer vision and motion planning. Robot could dynamically plan its action based on its perception of the environment. Even the integration of perception and planning are very promising in navigation and especially autonomous driving, they have limited impact on manipulation tasks due to tight precision tolerances in grasping and manipulation and relatively large uncertainty in perception. Moreover, due to the interactive nature of many manipulation tasks, a tiny uncertainty could cause unpredictable behaviors during a physical interaction and quickly lead to unrecoverable failure. 

The physical interaction behavior of a robot with the environment could be modeled as a dynamic system. Since many variables could only be estimated with uncertainty, and the actions have to be decided based on imperfect observations, the dynamic system has stochastic components. Markov process or related state-based approaches have been applied to the robot manipulation problem and similar problems with some success. Learning-based approaches, such as Hidden Markov Model (HMM) learning and reinforcement learning have been utilized in extracting human knowledge and manipulation strategies from their demonstrations \cite{Paulius2018Survey}. However, several important but difficult issues remain unsolved. 

This paper introduces our effort in addressing three pressing problems in service robotics: knowledge representation, motion learning generation, and grasping for manipulation. 

\section{Knowledge Representation}
\begin{figure}[t]
	\centering
	\includegraphics[width=8cm]{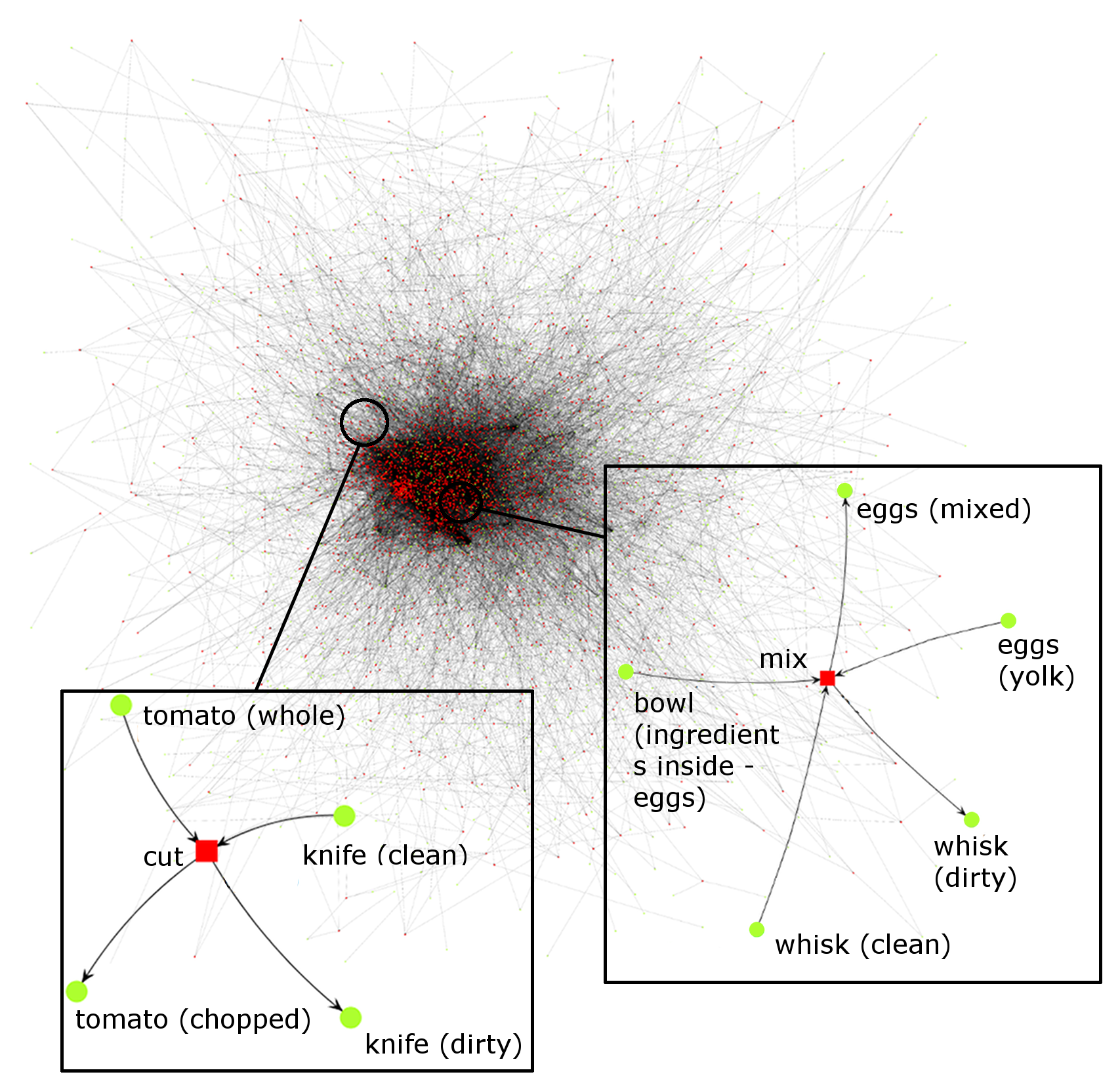}
	\caption{A visualization of our FOON combining knowledge from 65 instructional videos.
    Graphs were constructed for each video and merged into a single network to compress overlapping functional units and nodes.}
	\label{fig:foon}
\end{figure}

We are working on an approach that enables robots to learn skills by ``watching'' online instructional videos such as the ones on Youtube. From the video, we construct a functional object-oriented network (FOON) to represent manipulation knowledge (Figure \ref{fig:foon}) \cite{paulius2016functional,Paulius2017Learning,SunRAS2013}. The FOON is a graphical model that connects functional object motions with object to explicitly express the relationship between the functional-related objects \cite{Ren2013,sun2015modeling} and their functional motions in manipulation tasks. Combining the network structure and a recurrent neural network, we developed a novel video understanding approach that deciphers a cooking video into a semantic knowledge network that could be shared by all robots. Using the learned FOON, robots will be able to expand a task goal into a task tree, seek the correct objects at the desired states on which to operate, and generate optimal motion trajectories that are constructed from learned motion harmonics and satisfy novel environment and task constraints.

To extract cooking knowledge from online instructional videos, we have developed a four-stage video understanding pipeline to simultaneously recognize all events in a video and inferring the entire task in the video. The pipeline utilizes both objects and motions in the scene to identify events and tasks using a graph-based knowledge representation network as reference. Two sets of deep learning networks are trained to achieve this: 1) a convolutional neural network, which is deployed to recognize objects and to identify their positions, and 2) a recurrent neural network to understand the type of motion happening in the video. The closeness of objects to the human hand and the optical flow between frames is calculated to distinguish objects that are mainly involved in the event. These objects are referred to as objects-in-action. Confidence weights are assigned to objects based on their involvement in the scene and to motion classes based on their relatedness to the event happening using the results from the deep neural network. With a knowledge representation network FOON, confidence scores are computed to infer separate events in a video and understand the entire task. 

We performed experiments using our universal FOON (comprising of knowledge from 337 instructional videos) to analyze how FOON can identify events, incorporate history of events to recognize the flow of events and eventually how it can help to guess the recipe. Results show that using FOON produces a fair accuracy in recognizing events, and by fusing other features like automatically recognizing motion in video, we find an improvement in performance for event recognition. We also observed that recognizing objects-in-action influences the results of Functional Unit Recognition to a high extent. Therefore, in future work, we would like to explore other methods of identifying objects-in-action with higher accuracy since it is easier to infer the context of the activity by identifying the objects correctly \cite{Babeian2018}.

\section{Motion Generation}
One important component in the FOON is functional motion. Each functional motion associated to a daily-living tool has its own characteristic. For example, a whisk would be in a circular motion when it is used in beating eggs. The beating motion has a unique spatial-temporal pattern and is also bounded by the container of the eggs. We are developing an approach that learns the spatial-temporal motion patterns from human demonstrations using recurrent neural networks and then generates motion trajectories that can adapt to different constraints such as robot's kinematic/dynamic constraints, and environment constraints. 

Each time a daily manipulation is performed by humans, its execution is adjusted according to the environment and
is different from last time. To make robots more widely useful, researchers have been trying to help robots learn
a task and generalize to different situations, to which the approach of teaching robots by providing examples has
received considerable attention, known as programming by demonstration (PbD) \cite{billard2008}.

There are many existing motion trajectory generation frameworks. A popular one is called dynamical movement primitives (DMP) \cite{ijspeert_etal2013}. DMP is a stable non-linear dynamical system, and is capable of modeling discrete movement such as swinging a tennis racket \cite{ijspeert_etal2002}, playing table tennis  \cite{kober_etal2010} as well as rhythmic movement such as drumming \cite{schaal2003} and walking \cite{nakanishi2004}. DMP consists of a non-linear forcing function, a canonical system and a transformation system. The forcing function defines the desired task trajectory. 

To learn from human motions in interactive manipulation tasks, we have collected motion data, interactive force and torque data, environment data of 36 daily-living manipulation tasks in 3000 trials of 20 participants \cite{huang2018}, after reviewing all related existing datasets \cite{huang2016recent,huang2016datasets}. The tasks are mainly in food preparation, basic housework and maintenance, and personal hygiene. 

\begin{figure}[t]
	\centering
	\includegraphics[width=8cm]{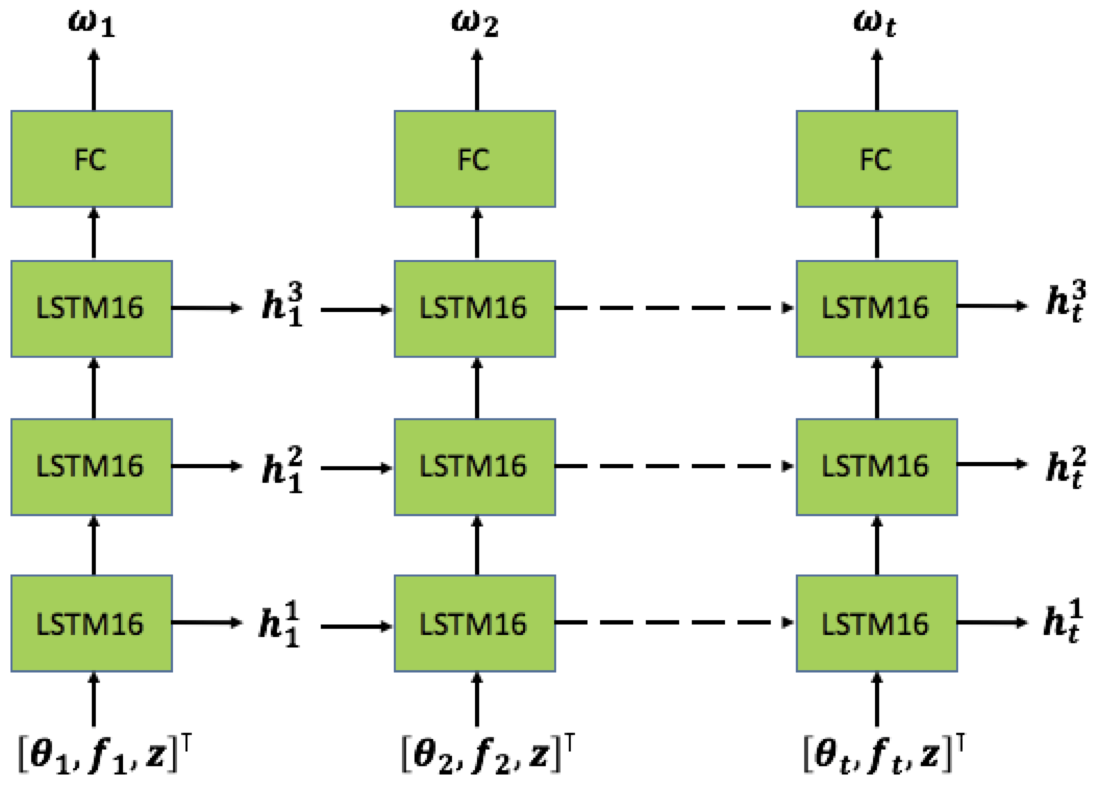}
	\caption{A Deep Recurrent Neural Network that could generate pouring motion based on the force reading in the cup's handle. The inputs to the network are force, current cup rotation angle, and cup parameters, while the out pout is the rotation angular velocity. }
	\label{fig:pouring}
\end{figure}

In our dataset, we selected pouring as one example of our motion generation application since pouring is a simple task people perform daily and it is the second most frequently executed motion in cooking scenarios, after pick-and-place \cite{paulius2016functional}. Using the collected data, we have developed a deep recurrent neural network \cite{graves2013} based approach to generate pouring trajectories, as illustrated in Figure \ref{fig:pouring}. The approach uses force feedback from the cup to determine the future velocity of pouring. We aim to make the system generalize its learned knowledge to unseen situations. In our evaluation experiments, the system successfully generalize when either a cup, a container, or the material changes, and starts to stumble when changes of more than one element are present \cite{Huang2017}. It is more flexible than our previous approach that applied functional principal component analysis \cite{huang2015generating}.

\section{Grasping for Manipulation}
We have noticed that there are extensive physical interactions between the tool we manipulate and the environment in many daily-living manipulation tasks \cite{huang2018}. Traditional pick-and-place grasps cannot accommodate the complex motion and force requirements in the manipulation tasks. To facilitate interactive manipulation motions, robotic grasps should efficiently transfer arm motion to manipulation motion and provide required force and torque. We have developed two novel grasp quality measures \cite{lin2016task, sun2016robotic, lin2015grasp} and an optimization approach to generate a grasp that can withstand and provide the interactive force/torque on the tool and efficiently facilitate manipulation motion during the task \cite{lin2015task, lin2015robot, lin2012learning}. 

\begin{figure}[t]
	\centering
	\includegraphics[width=8cm]{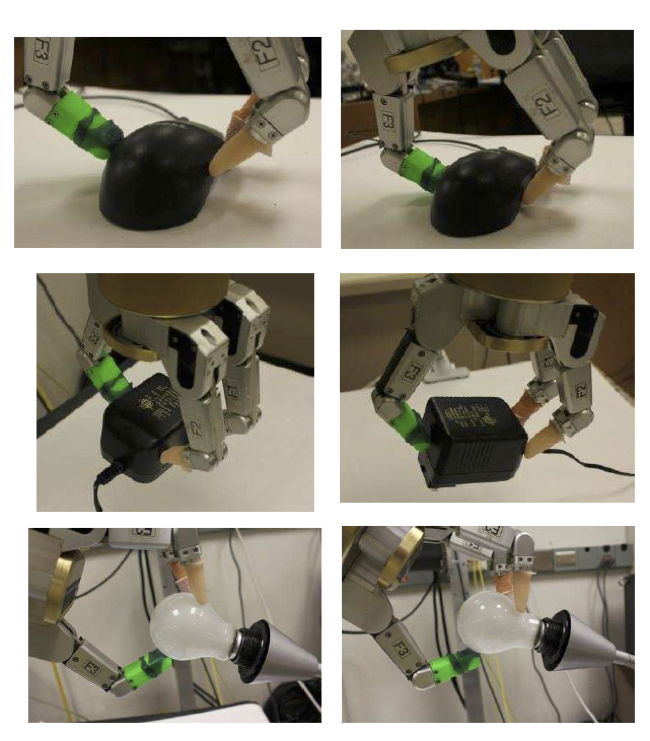}
	\caption{Evaluation of simulation on the real robotic platform.
Left column: execution result from our approach. Right column:
execution result from force-closure-based approach}
	\label{fig:grasps}
\end{figure}

We focuses on the grasp requirements derived from the voluntary and involuntary physical interactions in object manipulations. The manipulation-oriented grasp requirements include interactive wrench requirements and motion requirements that are required to accomplish a manipulation task. The manipulation-oriented grasp requirements are directly associated with the functionality of the instrument and the manipulation task, but independent from the robotic hardware. Grasp quality measures developed from the manipulation-oriented grasp requirements can be used as search criteria for optimal grasps. Working with hardware-independent grasping strategies extracted from human demonstration including grasp type and thumb placement, optimal grasps could be located efficiently in a dramatically reduced hand configuration space. Figure \ref{fig:grasps} shows the comparison of grasps in three typical physically interactive manipulation tasks: grasp generated using our approach vs. grasp generated using traditional force-closure-based approach. 

To facilitate the physical interaction in an tool manipulation task, robotic grasp should meet the interactive wrench and motion demands that are required to accomplish the manipulation. Those demands are independent from the robotic hardware, but are directly derived from the functionality of the tool and the manipulation task and represent the dynamics of the interaction. Two grasp quality measures are developed from the interactive wrench and motion requirements and then used as search criteria for optimal grasps. The search or optimizing process can be dramatically improved by narrowing down the optimization search space using prior grasping strategies that are independent from robotic hardware, including grasp type and thumb placement.

\section{Discussion and Future Work}
Our research in the presented directions are on-going efforts toward robot cooking. In cooking knowledge extraction, we are working on automatically recognizing objects and finding their bounding boxes. However, this would require much more training data and cycles to fine-tune the performance for deep neural network approaches. We also wish to generalize the knowledge contained within a FOON so that we could achieve better results in inference and solving the problem of event history incorporation. 

For physical-interactive manipulation motion generation, we plan to develop a robust learning approach that learns from the demonstration data in the form of a broader representation rather than an optimized but narrowed fitting, so that it is easy to incorporate the learning from demonstration into a model-predictive control approach during run-time that could adapt to unseen situations. After pouring, we plan to apply our approach to other commonly performed manipulations for cooking such as cutting, stirring, and others. For grasping, we plan to model the required motion and wrench in the manipulations using statistics models and then generalize those models to unobserved manipulations so that a limited number for human demonstration could be sufficient. In the end, we plan to combine all the components together to develop a comprehensive robot cooking solution for our daily-living environment. 

\begin{acks}

This material is based upon work supported by the National
Science Foundation under Grants No. 1421418 and
No. 1560761.

\end{acks}

\bibliographystyle{ACM-Reference-Format}
\bibliography{sample-bibliography}

\end{document}